\documentclass[10pt,twocolumn,letterpaper]{article}
\usepackage[pagenumbers]{cvpr}
\def\paperTitle{CoHABIT: Complementary Evaluation of Furniture Layouts\\through Spatial Rules and Household Routines}

\usepackage[T1]{fontenc}

\usepackage{DejaVuSans}
\usepackage{microtype}
\usepackage{capt-of}
\usepackage{tabularx,array,longtable,ragged 2e}
\usepackage{enumitem}
\usepackage[most]{tcolorbox}
\usetikzlibrary{positioning,calc,arrows.meta}
\definecolor{preferencepurple}{HTML}{75549F}
\usepackage{placeins}
\usepackage{etoolbox}
\definecolor{cvprblue}{HTML}{2166AC}
\definecolor{staticblue}{HTML}{276A9D}
\definecolor{dynamicorange}{HTML}{A96123}
\definecolor{successgreen}{HTML}{267568}
\definecolor{failurered}{HTML}{AD3E49}
\definecolor{ink}{HTML}{213745}
\definecolor{rulegray}{HTML}{D3DCE2}
\newcolumntype{Y}{>{\RaggedRight\arraybackslash}X}
\newcolumntype{P}[1]{>{\RaggedRight\arraybackslash}p{#1}}

\newcommand{\suppsec}[1]{Supplement, Sec.~\csname suppletter@#1\endcsname}

\tcbset{figurepanel/.style={enhanced,colback=white,colframe=rulegray,boxrule=.55pt,
  arc=1.5pt,left=6pt,right=6pt,top=5pt,bottom=5pt,before skip=0pt,after skip=0pt}}
\setlist[itemize]{topsep=3pt,itemsep=1pt,parsep=0pt,leftmargin=*}
\setlist[enumerate]{topsep=3pt,itemsep=1pt,parsep=0pt,leftmargin=*}
\expandafter\def\csname suppletter@evidence_boundaries\endcsname{A}
\expandafter\def\csname suppletter@reporting_model\endcsname{A}
\expandafter\def\csname suppletter@static_track\endcsname{B}
\expandafter\def\csname suppletter@hanfu\endcsname{B}
\expandafter\def\csname suppletter@dynamic_model\endcsname{C}
\expandafter\def\csname suppletter@operational_model\endcsname{C}
\expandafter\def\csname suppletter@learning_details\endcsname{C}
\expandafter\def\csname suppletter@resident_suitability\endcsname{D}
\expandafter\def\csname suppletter@complementarity_proof\endcsname{D}
\expandafter\def\csname suppletter@human_protocol\endcsname{D}
\expandafter\def\csname suppletter@ablation_matrix\endcsname{D}
\expandafter\def\csname suppletter@construction_audit\endcsname{E}
\expandafter\def\csname suppletter@whole_apartment_semantics\endcsname{E}
\expandafter\def\csname suppletter@behavioral_cloning\endcsname{F}
\expandafter\def\csname suppletter@new_results_v 7\endcsname{F}
\expandafter\def\csname suppletter@matched_fixture\endcsname{F}
\expandafter\def\csname suppletter@two_room_trace\endcsname{F}
\expandafter\def\csname suppletter@policy_diagnostics\endcsname{F}
\expandafter\def\csname suppletter@reproducibility\endcsname{G}
\expandafter\def\csname suppletter@submission_boundary\endcsname{G}

\expandafter\def\csname suppletter@frozen_pair\endcsname{F}

\expandafter\def\csname suppletter@architecture_learning\endcsname{C}
\expandafter\def\csname suppletter@criterion_algorithms\endcsname{C}
\expandafter\def\csname suppletter@static_methods\endcsname{B}

\usepackage[pagebackref,breaklinks,colorlinks,allcolors=cvprblue]{hyperref}
\hypersetup{pdftitle={CoHABIT: Complementary Evaluation of Furniture Layouts through Spatial Rules and Household Routines},pdfauthor={Ewsey Obzherin; Samira Atlangirieva; Anastasia Antsiferova; Valentin Khrulkov; Art Prosvetov; Andrey Kuznetsov}}
\title{\paperTitle}
\author{%
Ewsey Obzherin\textsuperscript{1,2} \quad Samira Atlangirieva\textsuperscript{1,3} \quad Anastasia Antsiferova\textsuperscript{1,4}\\
Valentin Khrulkov\textsuperscript{1} \quad Art Prosvetov\textsuperscript{1} \quad Andrey Kuznetsov\textsuperscript{1,4}\\[4pt]
{\normalsize\textsuperscript{1}FusionBrain Lab}\\
{\normalsize\textsuperscript{2}Moscow Independent Research Institute of Artificial Intelligence}\\
{\normalsize\textsuperscript{3}Higher School of Economics \qquad \textsuperscript{4}Innopolis University}
}

\makeatletter
\patchcmd{\@maketitle}{\vskip .375in}{\vskip .125in}{}{\PackageError{cohabit}{Title top-spacing patch failed}{}}
\patchcmd{\@maketitle}{\vspace*{12pt}}{\vspace*{0pt}}{}{\PackageError{cohabit}{Title bottom-spacing patch failed}{}}
\patchcmd{\@maketitle}{\vspace*{24pt}}{\vspace*{12pt}}{}{\PackageError{cohabit}{Title author-spacing patch failed}{}}
\apptocmd{\@maketitle}{
\begin{center}
\includegraphics[width=\textwidth,height=2.05in,keepaspectratio]{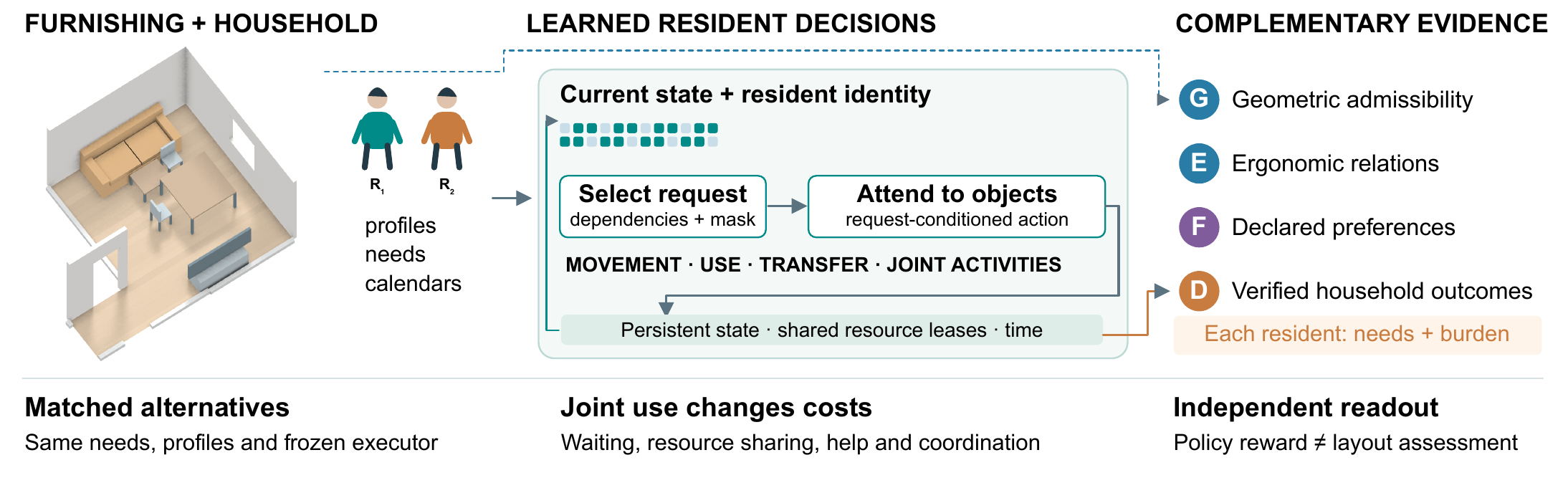}
\captionof{figure}{\textbf{CoHABIT evaluates furnishings through complementary evidence.} A source-geometry room and schematic residents connect spatial criteria to learned household execution. Residents share actor weights but make separate decisions in a persistent world; independent need and burden readouts support matched comparisons.}
\label{fig:architecture-overview}
\end{center}

}{}{}
\makeatother
\begin{document}
\raggedbottom

\setcounter{dbltopnumber}{1}
\renewcommand{\dbltopfraction}{.75}
\renewcommand{\textfraction}{.02}
\renewcommand{\dblfloatpagefraction}{1}
\maketitle
\begin{abstract}
We introduce \textbf{CoHABIT}, a framework for evaluating furniture layouts through complementary spatial rules and household routines. Static assessment captures geometric validity, ergonomic access, functional organization, and configurable design preferences, including Feng Shui. Dynamic assessment uses learned resident policies to simulate everyday activities, resource sharing, coordination, and individual burdens in multi-agent households. Controlled comparisons show how spatial criteria and household behavior reveal distinct aspects of layout quality, while resident profiles and scheduling influence furnishing rankings. Independent verification of household needs grounds dynamic comparisons in observed outcomes, enabling interpretable assessments of how furnished environments serve different households.
\end{abstract}

\section{Introduction}
\label{sec:intro}
\begin{figure*}[t]\centering
\includegraphics[width=\textwidth]{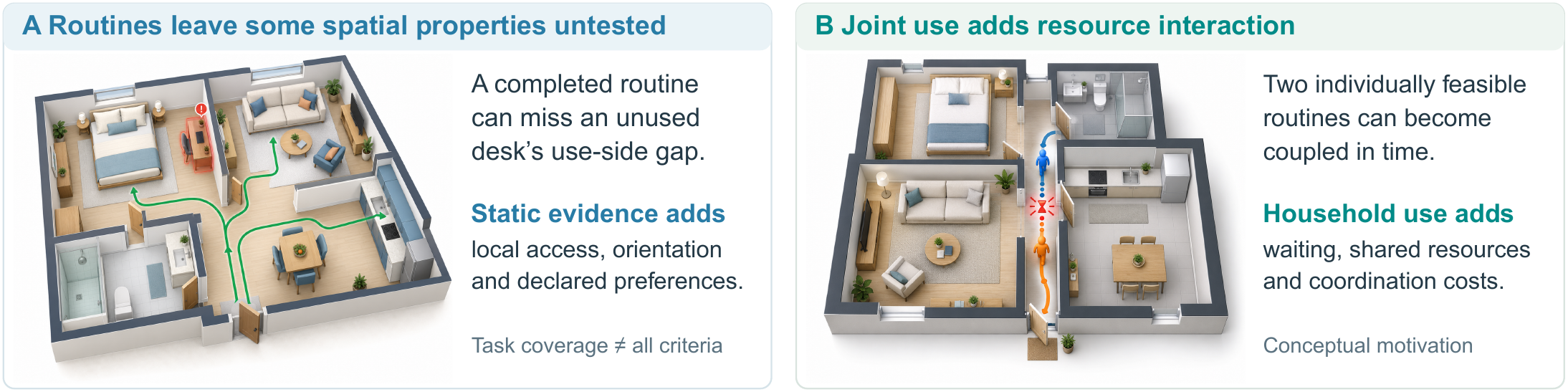}
\caption{\textbf{Why spatial criteria and routines complement each other.} A: a completed routine can leave an unused object's access untested. B: individually feasible routines can become coupled during joint use. These AI-generated conceptual illustrations motivate task coverage and concurrent use; arrows are schematic.}
\label{fig:motivation}
\end{figure*}

Evaluating furnished apartments requires more than plausible appearance or collision-free placement. A reachable desk may lack chair clearance; an accessible bed may violate an orientation preference; a shared kitchen may impose waiting. These examples motivate complementary spatial and household evidence (Fig.~\ref{fig:motivation}).

Co-Layout~\cite{xiang2026colayout} synthesizes coherent room--furniture arrangements; Function2Scene and SceneTeract~\cite{wang2026function2scene,maillard2026sceneteract} encode occupant requirements or verify action feasibility. We ask which alternative better serves a declared household through a shared routine, and why.

During a family evening, one adult prepares dinner while another helps a child before everyone shares a meal. Birthday preparation adds shared activities and cleanup. The same furniture supports changing roles and competing needs. We represent quality as a household-conditioned vector: spatial requirements, declared preferences and consequences of use.

We test when one perspective distinguishes alternatives another leaves tied. \emph{Furnishing interventions} fix envelope, inventory, residents, original needs and starts while changing placement or orientation. \emph{Apartment comparisons} fix household, semantic needs, calendar and policy across homes with admitted furniture bindings. Independent verification checks need fulfillment before costs; personal outcomes preserve tradeoffs and incomplete execution remains unresolved.

Our contributions are \textbf{multi-perspective assessment}, retaining interpretable geometric, ergonomic, preference and household measurements; \textbf{household-conditioned temporal assessment}, comparing shared resources, assistance and dependent routines through personal costs; and \textbf{competence-aware learned evaluation}, checking original needs independently of the executor. Controlled interventions and audited trajectories show agreement, resolved static ties and household-dependent tradeoffs under reproducible matched inputs.

\section{Related Work}
\paragraph{Spatial and semantic assessment.}
Design guidelines and human-centric metrics assess geometric access and functional relationships~\cite{merrell2011,yu2011,neufert2019,fu2020human,yang2024physcene}. SceneEval separates explicit requirements from implicit plausibility~\cite{tam2026sceneeval}; SceneCritic diagnoses semantic, orientation and geometric relations~\cite{sengupta2026scenecritic}. LayoutVLM, DisCo-Layout, AutoLayout and Agentic Designer use such feedback in refinement~\cite{layoutvlm2025,gao2025discolayout,chen2025autolayout,yang2026agenticdesigner}. Scene-Designer, contemporaneous work, learns a scene critic and uses it in GRPO training and at test time~\cite{liu2026scenedesigner}. Its critic provides scene-level judgments; CoHABIT conditions assessment on a declared household and reports spatial criteria alongside resident-execution outcomes. Cultural and Feng Shui preferences form an explicit configurable rubric~\cite{ham2004crosscultural,mak2005fengshui,hong2016fengshui}.

\paragraph{Activity- and human-aware layouts.}
Human-action priors inform scene synthesis~\cite{savva2014,fisher2015,mover2022,mime2023}. Function2Scene converts occupant/activity specifications into constraints for generation and refinement~\cite{wang2026function2scene}. SceneTeract combines VLM task decomposition with agent-specific geometric feasibility checks; verifier feedback additionally trains VLM feasibility predictions with GRPO~\cite{maillard2026sceneteract}. Its semantic/geometric complementarity concerns whether an activity can be executed. Behavior-Aware Anthropometric Scene Generation reserves manipulation space and studies individual and collaborative use, including six three-person teams~\cite{jin2026behavior}. CoHABIT studies a complementary comparison target: consequences of dependent needs in a persistent apartment shared by a declared household. Matched calendars, resources and assistance expose total and personal costs, alongside geometry, ergonomics and preferences. The contribution is this combined, household-conditioned assessment formulation.

\paragraph{Generation and simulation environments.}
Scene generators produce furnished alternatives~\cite{atiss2021,diffuscene2024,layoutgpt2023,holodeck2024,xiang2026colayout}; SAGE combines critics for environment generation~\cite{xia2026sage}. ProcTHOR, SceneSmith and HomeWorld supply simulation-oriented scenes~\cite{deitke2022procthor,pfaff2026scenesmith,li2026homeworld}. Household simulators support stateful activities~\cite{virtualhome2018,alfred2020,behavior2021,li2024behavior,igibson2021}, while Watch-And-Help, Habitat and MAPPO address cooperation~\cite{wah2021,habitat2024,yu2022mappo}. We use repeatable comparisons with fixed needs and policies. Independent need verification separates executor competence from furnishing evidence: incomplete policy runs retain an unresolved furnishing assessment.

\section{Problem Formulation}
\label{sec:framework}
Given a furnished layout $\mathcal{L}$, household $H$ and declared preference rubric $\rho$, we seek an interpretable assessment
\begin{equation}
\mathcal Q(\mathcal{L},H,\rho)=\bigl(G(\mathcal{L}),E(\mathcal{L},H),F_\rho(\mathcal{L}),D_H(\mathcal{L})\bigr).
\label{eq:quality-vector}
\end{equation}
$G$ records geometric validity; $E$ records functional and ergonomic criteria, conditioned on resident capabilities where required; $F_\rho$ records the active spatial preferences; and $D_H$ records need fulfillment, burden and interactions during use. Each component retains its measurements, applicability and missingness for separate interpretation. Inputs specify metric geometry, facing and affordances, household capabilities, and original needs with resources and time windows.

\begin{figure*}[t]\centering
\begin{minipage}[t]{.49\textwidth}\centering\sffamily\fontsize{8}{10}\selectfont
\textbf{G: geometry --- furniture near an opening}\par\smallskip
{\fontsize{8}{9.5}\selectfont\makebox[.5\linewidth][c]{A}\makebox[.5\linewidth][c]{B}}\par
\includegraphics[width=\linewidth]{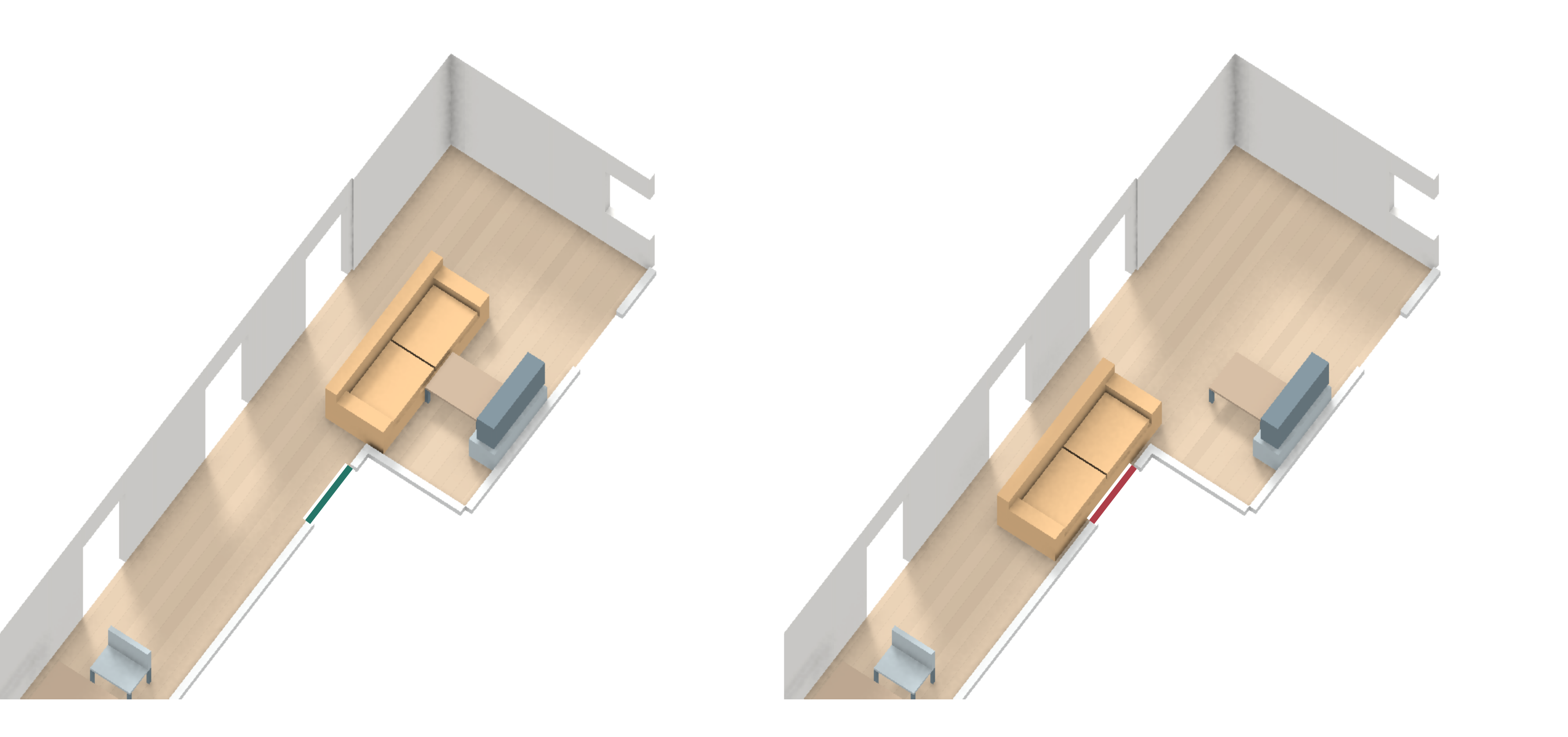}\par
{\small Opening-related findings: $0\rightarrow3$.}
\end{minipage}\hfill
\begin{minipage}[t]{.49\textwidth}\centering\sffamily\fontsize{8}{10}\selectfont
\textbf{E: ergonomics --- the working side matters}\par\smallskip
{\fontsize{8}{9.5}\selectfont\makebox[.5\linewidth][c]{A}\makebox[.5\linewidth][c]{B}}\par
\includegraphics[width=\linewidth]{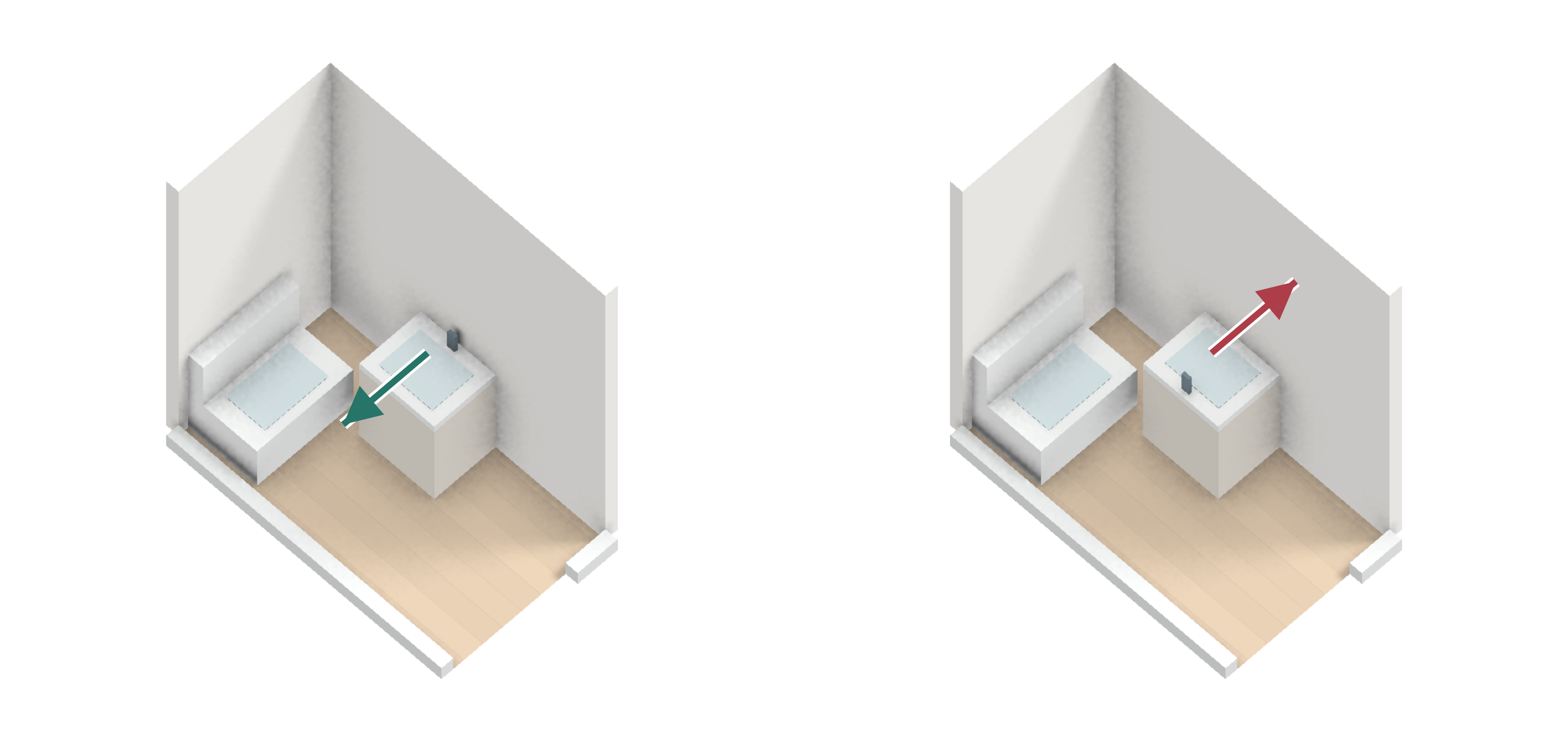}\par
{\small Same footprint; the sink faces the wall in B.}
\end{minipage}
\par\medskip
\begin{minipage}[t]{.49\textwidth}\centering\sffamily\fontsize{8}{10}\selectfont
\textbf{F: preferences --- headboard support}\par\smallskip
{\fontsize{8}{9.5}\selectfont\makebox[.5\linewidth][c]{A}\makebox[.5\linewidth][c]{B}}\par
\includegraphics[width=\linewidth]{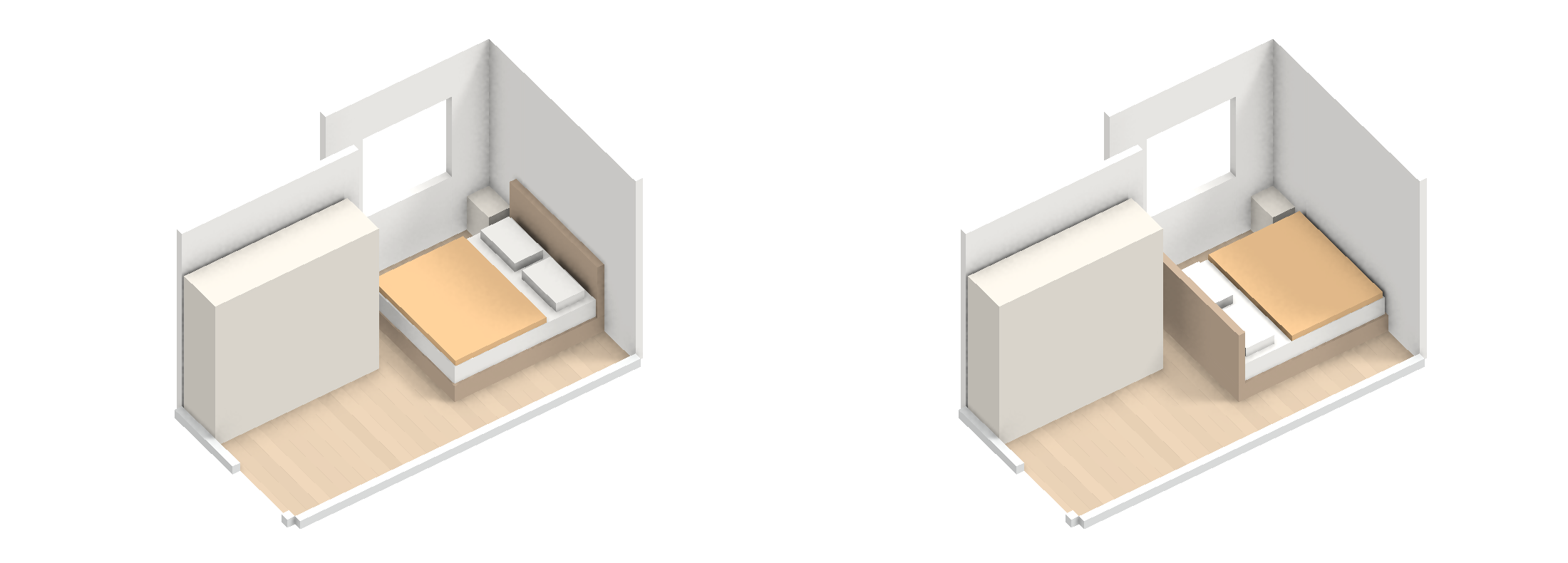}\par
{\small Backing contact is lost after rotation.}
\end{minipage}\hfill
\begin{minipage}[t]{.49\textwidth}\centering\sffamily\fontsize{8}{10}\selectfont
\textbf{D: dynamics --- furnishing order depends on $H$}\par\smallskip
{\fontsize{8}{9.5}\selectfont\makebox[.5\linewidth][c]{A}\makebox[.5\linewidth][c]{B}}\par
\includegraphics[width=\linewidth]{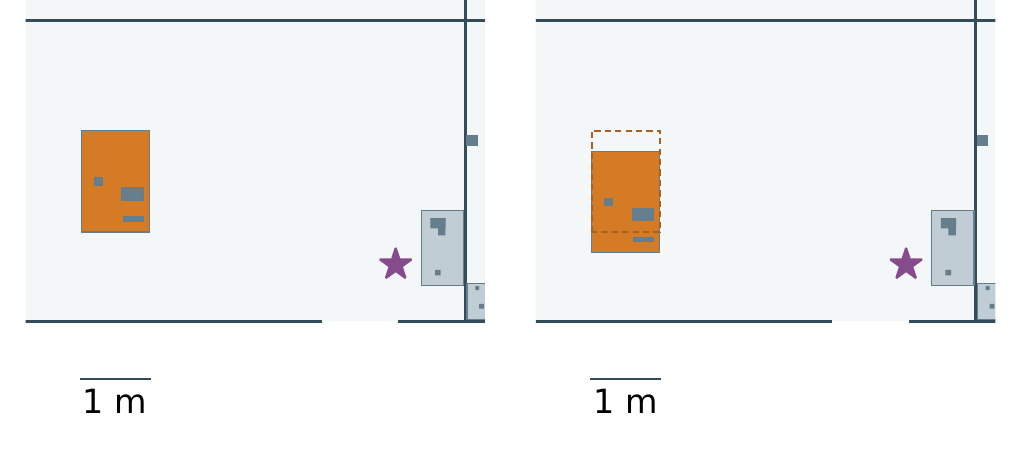}\par
{\small The preferred furnishing reverses with the profile.}
\end{minipage}
\caption{\textbf{Paired examples of complementary criteria.} G: opening and clearance (colored opening). E: facing-dependent access (arrows). F: headboard support. 3D views preserve source footprints and facing; heights/materials are schematic. D: profile-dependent cost after a table move, using complete frozen-policy executions. Each pair preserves its envelope and inventory.}
\label{fig:critic-pairs}
\end{figure*}

\textbf{Diagnostic complementarity.} For applicable observed criteria $Q_i,Q_j$ and assessment instances $a=(\mathcal{L}_a,H_a,\rho_a)$ and $b=(\mathcal{L}_b,H_b,\rho_b)$, a local witness is $Q_i(a)=Q_i(b)$ but $Q_j(a)\ne Q_j(b)$. Criterion $j$ supplies a criterion-level distinction left unresolved by $i$. In Figure~\ref{fig:critic-pairs}E--F, footprint-based geometric checks stay unchanged while use-side access or the declared headboard-support criterion changes.

\section{Static and Preference-Based Assessment}
\label{sec:static}
\paragraph{Geometry and ergonomics.}
Figure~\ref{fig:critic-pairs}G--E contrasts an opening obstruction with changed use-side access at an unchanged footprint.
We separate required geometric checks from recommendations. Containment, unauthorized overlap and required access have explicit tolerances and contact semantics; permitted support contacts are excluded from overlap violations. Use-side clearance, orientation and functional relationships are measured for the specified resident. Each criterion records its units, source, applicability and availability. A hard PASS requires every declared required check to be available and satisfied; missing evidence gives UNKNOWN unless a verified failure already applies.

The training-free geometry/ergonomics critic ranks layouts by $3n_{\rm critical}+n_{\rm advisory}$, lower being better. Here ``critical denotes implementation severity, covering hard checks (collision, bounds, door and window zones) and recommendations such as gaps; Eq.~(S2) defines the required-check set separately. This score is used in the static perturbation experiment. The VLM judge provides separate pair judgments from plan images. A static scene critic evaluates an arrangement; the centralized RL value critic estimates execution return. They serve different roles.

\paragraph{Feng Shui: optional, versioned preferences.}
Figure~\ref{fig:critic-pairs}F isolates the headboard-support subrule: identical bed footprints can satisfy or violate its backing requirement.
HANFU is CoHABIT's household-conditioned spatial-preference rubric. Its ten criteria cover commanding positions for bed/desk/sofa, Bagua, kitchen positioning, directed corners, door--bed alignment, mirrors, elemental balance and rest/activity zoning. These cultural preferences remain separate from geometric admissibility and ergonomic advice. A commanding-position judgment needs facing, entrance and backing geometry; directional rules need compass bearing and a declared sector construction. Compass orientation is supplied explicitly. Material- and mirror-based criteria need the corresponding scene attributes.

Reports retain criterion values, the active set and applicability. Appendix~B specifies HANFU, its inputs, and the criterion definitions used in our preference experiments.

\section{Dynamic Household Assessment}
\label{sec:dynamic}
Figure~\ref{fig:architecture-overview} separates decisions, shared execution and independent assessment. Figure~\ref{fig:critic-pairs}D changes furniture in one home; Figure~\ref{fig:whole-home-pair} compares homes through the same routine. Residents pursue needs while actions change locations, ownership and resource availability. Each ready resident selects a request and action; completed actions update the persistent state.

At event $n$, $x_n$ is the shared state, $\mathcal I_n$ the ready residents, and $\xi$ the fixed request realization. With the same frozen actor across furnishings, decisions factor as
\begin{align}
\boldsymbol\pi_\theta(\mathbf q_n,\mathbf a_n\mid\mathbf o_n)
&=\prod_{i\in\mathcal I_n}\pi_\theta(q_n^i\mid o_n^i)
\widetilde\pi_\theta(a_n^i\mid o_n^i,q_n^i),\\
(x_{n+1},\Delta t_n)&\sim P_\mathcal{L}(\cdot\mid x_n,\mathbf q_n,\mathbf a_n,H,\xi).
\end{align}
The actor conditions on objects, profiles and released needs to select movement, manipulation, transfer, waiting or sustained activity. 

\paragraph{Scope of dynamic evidence.}
Dynamic execution models logical state transitions and geometry-aware assisted navigation; continuous body motion and mutual passage lie outside this model. Static checks $G,E$ measure clearances under declared profiles; execution measures temporal outcomes. Figure~S7 (Appendix~B) illustrates reduced chair clearance despite completed routines. Travel distance sums committed NAV-path lengths, including repeated traversals (Fig.~\ref{fig:whole-home-pair}).

\paragraph{Learning roles.}
MAPPO uses a joint-state value critic for advantages and decentralized resident decisions; GRPO forms a baseline from sampled completions of a policy~\cite{yu2022mappo,shao2024deepseekmath}. Grouped samples are distinct from interacting residents. Our actor--critic studies use MAPPO; later request-conditioned daily and family actors use behavioral cloning (BC). Appendix~C identifies both protocols. Comparisons fix weights, admissibility and decoding; changing a mask defines a different executed policy. Reference and temporal controllers are evaluated independently of learned actors.

\paragraph{Target of household assessment.}
For household $H=\{p_i\}_{i=1}^N$, let $\xi\sim\nu_H$ denote its exogenous needs. A frozen policy produces $\tau\sim P_\mathcal{L}^{\pi_\theta}(\cdot\mid H,\xi)$. The assessment target is
\begin{equation}
D_H(\mathcal{L})=\mathbb E_{\xi,\tau}[\mathbf B(\tau),\mathbf Y(\tau),\mathbf W(\tau)],
\end{equation}
where $\mathbf B$ records resident burdens, $\mathbf Y$ required, met, verified-unmet and unresolved needs, and $\mathbf W$ attributable interactions. These outcomes define the external assessment, with training objectives specified separately. Primary burden includes instrumental time, invalid attempts and witnessed forced waiting, excluding desired eating, rest and work. Voluntary waiting is a sensitivity control. Household cost $C_\Sigma=\sum_i B_i$ is compared on complete matched executions.

Profiles separate capabilities, request distributions, burden weights and behavior. Interventions vary mobility, synchrony, kitchen frequency or permitted assistance with other inputs fixed. Assistance changes dispatch (Appendix~D). Complete matched joint and solo executions isolate shared use:
\begin{equation}
I(\mathcal{L},H)=C_{\rm joint}(\mathcal{L},H)-\sum_i C_{\rm solo}(\mathcal{L},p_i).
\label{eq:interaction-cost}
\end{equation}
Needs, starts, cost definition and policy are fixed; joint-only obligations are evaluated through a separate intervention. Event records attribute changes in $I$ to contention, waiting, assistance and joint actions. Family and furnishing examples additionally report pending-need latency, $L_\Sigma=\sum_i\int_0^T N_i^{\rm pending}(t)\,dt$, integrating outstanding original needs over time. It differs from instrumental burden $C_\Sigma$. Each experiment declares its cost functional before comparing alternatives.

\subsection{Evidence, costs and three-state verdicts}
\label{sec:joint}
The readout also retains path length, active and waiting time, deadline misses and conflicts. Unresolved needs remain visible, and coverage records the observed needs and actions. A positive requires the original AND-needs and declared cost limits to be verified. A negative needs an independent impossibility certificate or justified cost attribution under competent execution. Unresolved policy, source or runtime failure remains unknown.

A conservative feasibility gate combines required spatial conditions AND verified need-fulfillment and cost conditions. A supported failure prevails, positive requires both known and passed, and otherwise the verdict is unknown. This feasibility gate and the complementary assessment vector in Eq.~\ref{eq:quality-vector} serve separate roles. Required failures determine the feasibility verdict independently of preference scores.

Appendix~D specifies the matched contrasts. Local witnesses and their trajectories establish the added information; the gate determines when a comparison is interpretable.

\section{Benchmarks and Experimental Protocol}
\label{sec:benchmark}
\paragraph{Static edits with controlled provenance.}
Fifty furnished controls yield 300 control--edit pairs: shifts, rotations, blocked doors, overlap, wall crossing and combinations. The 19 source layouts span 11 contour families; the 48 test pairs use eight controls from three layouts and two held-out contour families. Each control and its edits share one split. Controls satisfy selected checks. Pair accuracy gives half credit to ties; detection, localization and repair use separate denominators (Appendix~B). The cascade queries a VLM when the rule-score gap is below three; the ergonomics-first tandem consults preferences on critic ties.

\paragraph{Households and matched comparisons.}
\label{sec:resident-profiles}
Profiles specify clearance, speed, reach and assistance. The rich corpus uses standard--fast and standard--wide-clearance households; daily comparisons use standard--standard and standard--slow. Fast changes speed alone; wide-clearance and slow are compound interventions. The family adds a child with declared numerical capabilities. Appendix~A illustrates these simulator profiles and lists their parameters.

The rich and daily cohorts remain distinct. Daily comparisons use five development families admitted by geometry, common-start and task-binding checks. Each pair fixes household, original needs, starts and calendar; navigation is verified per layout (Appendix~E).

\subsection{Scope and limitations}
CoHABIT evaluates formal criteria and declared routines. Numerical profiles specify simulator capabilities. Appendix~B records criterion versions and evidence provenance; sample counts distinguish source families from repeated conditions.

\section{Experiments}
\label{sec:experiments}
\subsection{Complementary static criteria}
The rule critic outperforms collision and reachability baselines in synthetic pair ordering; the static--VLM cascade has the highest overall accuracy (Table~\ref{tab:static-ranking-main}). The critic, the VLM judge and cascade have comparable observed scores on the small 48-pair test; Appendix~B details uncertainty and edit-specific results.
\begin{table}[t]\centering\small
\setlength{\tabcolsep}{3pt}
\caption{Archived pair accuracy (\%). All: 300 pairs, 50 controls, 11 contour families, including training controls. Test: 48 pairs, eight controls, two families. Brackets in All: reported 95\% bootstrap CIs with controls as the resampling unit; --: unreported.}
\label{tab:static-ranking-main}
\begin{tabular}{@{}lrr@{}}\toprule
Method & All [95\% CI] & Test\\\midrule
Geometry/ergonomics &91.7 [88.8, 94.5]&95.8\\
SceneEval collision &77.0 [--]&77.1\\
SceneEval accessibility &66.7 [--]&62.5\\
SceneSmith reach &57.2 [--]&63.5\\
VLM judge &94.8 [92.7, 96.7]&94.8\\
Static--VLM cascade &98.3 [97.0, 99.3]&99.0\\\bottomrule
\end{tabular}\end{table}
\begin{figure*}[t]\centering
\includegraphics[width=\textwidth]{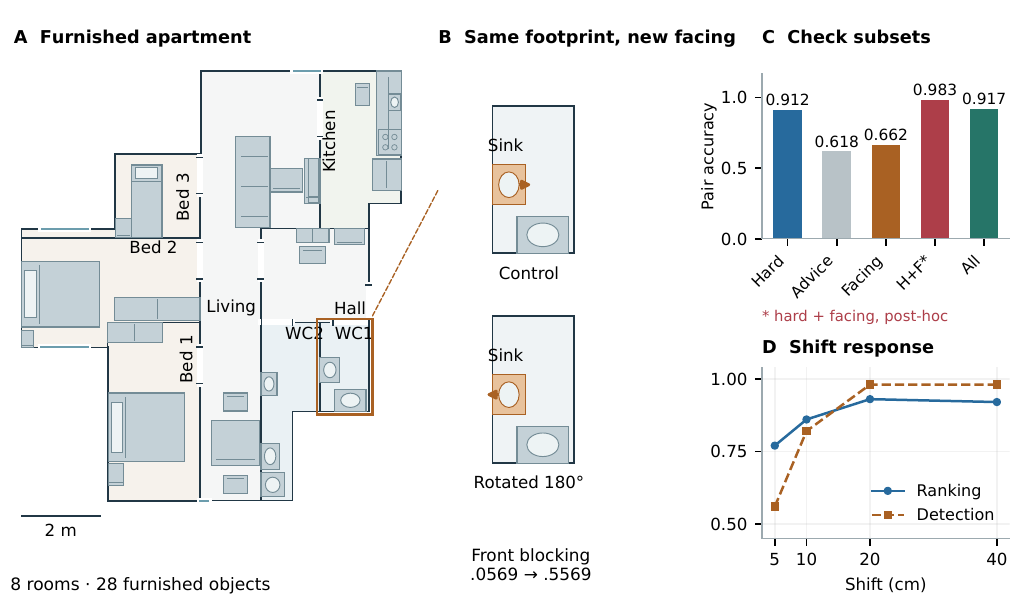}
\caption{\textbf{Static inspection from apartment to object.} A--B: an eight-room furnished plan and an orientation edit invisible to footprint checks. C: exploratory ablations (star: post-hoc selection). D: detection and ranking across displacement magnitudes.}
\label{fig:static-study}
\end{figure*}

\paragraph{Where rules complement visual judgments.}
On 50 rotations, the VLM judge scores 78\%, versus 95\% for rules and 96\% for its cascade. A preserved sink example exposes the mechanism: rotating the working side reduces the working-side distance to the wall from 0.75\,m to zero; rules detect it, while the judge ties the pair. Rules additionally identify offending objects and measured clearances. They run locally without API calls; the cascade queries a judge for about 20\% of pairs, reducing external requests by about 80\%. Appendix~B reports historical API charges and individual examples.

\paragraph{Orientation adds information beyond footprint checks.}
Rotation edits preserve overlap, opening and containment flags, yet reveal orientation-dependent use-side blocking (Fig.~\ref{fig:static-study}). Combining hard checks with orientation performs strongly in a post-hoc subset. Gap penalties expose a distinction between synthetic ordering and ergonomic advice. Increasing displacement improves detection more consistently than ranking. Localization and repair are measured through changed objects and removed findings (Appendix~B).

\begin{figure}[t]\centering
\includegraphics[width=\linewidth]{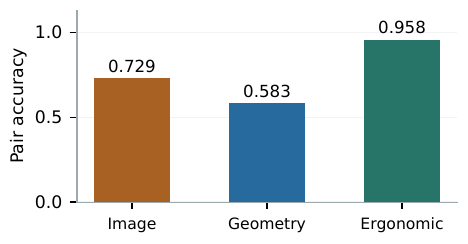}
\caption{\textbf{Preference pair ordering.} Five-criterion rubric (Appendix~B): results use its own criterion definitions, distinct from HANFU.}
\label{fig:preference-evidence}
\end{figure}

\begin{figure}[t]\centering
{\sffamily\fontsize{8}{9.5}\selectfont\setlength{\tabcolsep}{1pt}
\begin{tabular}{@{}cc@{}}
\textbf{Input} & \textbf{Selected}\\[-1pt]
\multicolumn{2}{c}{Living room: shift and rotate the sofa}\\[1pt]
\includegraphics[viewport=2 254 246 429,clip,width=.485\linewidth]{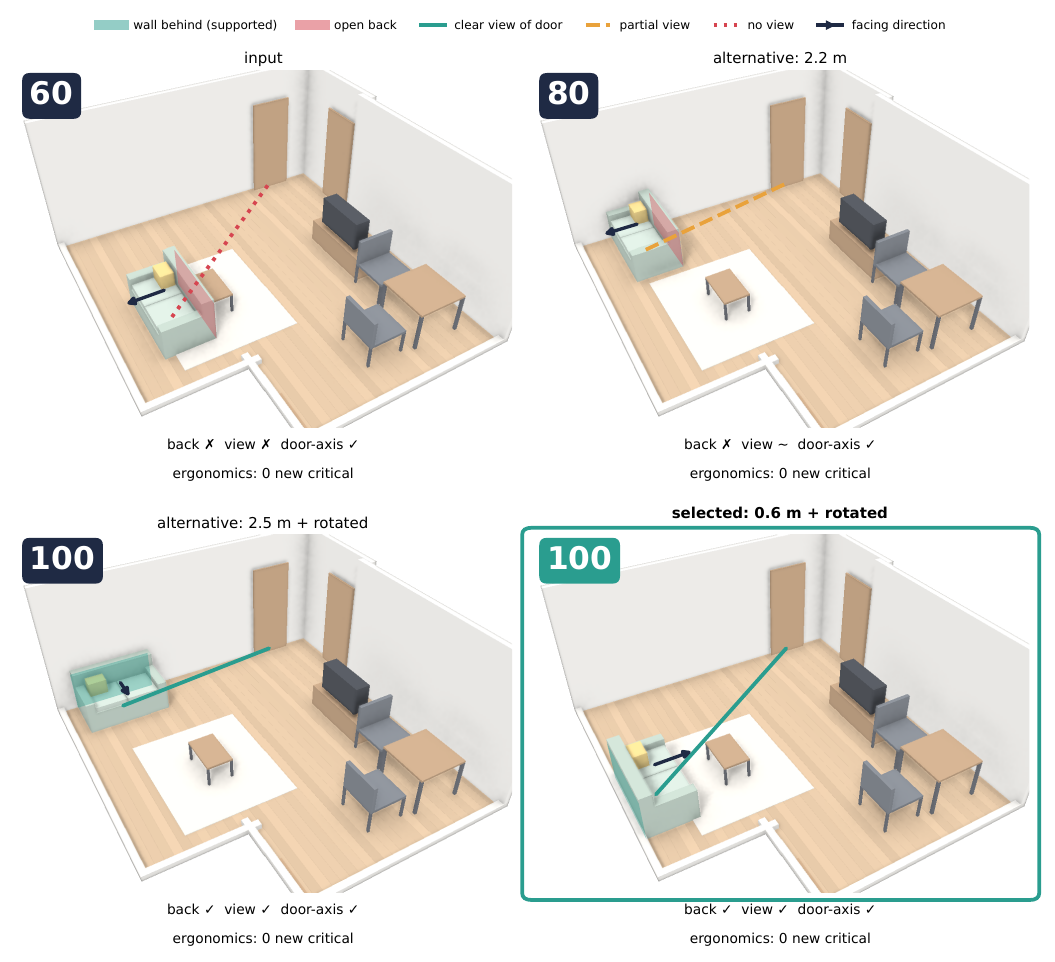}&
\includegraphics[viewport=249 29 505 211,clip,width=.485\linewidth]{figures/static_preference_living_archive.pdf}\\[-2pt]
Open back; no door view & Wall behind; clear view\\[3pt]
\multicolumn{2}{c}{Bedroom: rotate the bed}\\[1pt]
\includegraphics[viewport=2 254 246 429,clip,width=.485\linewidth]{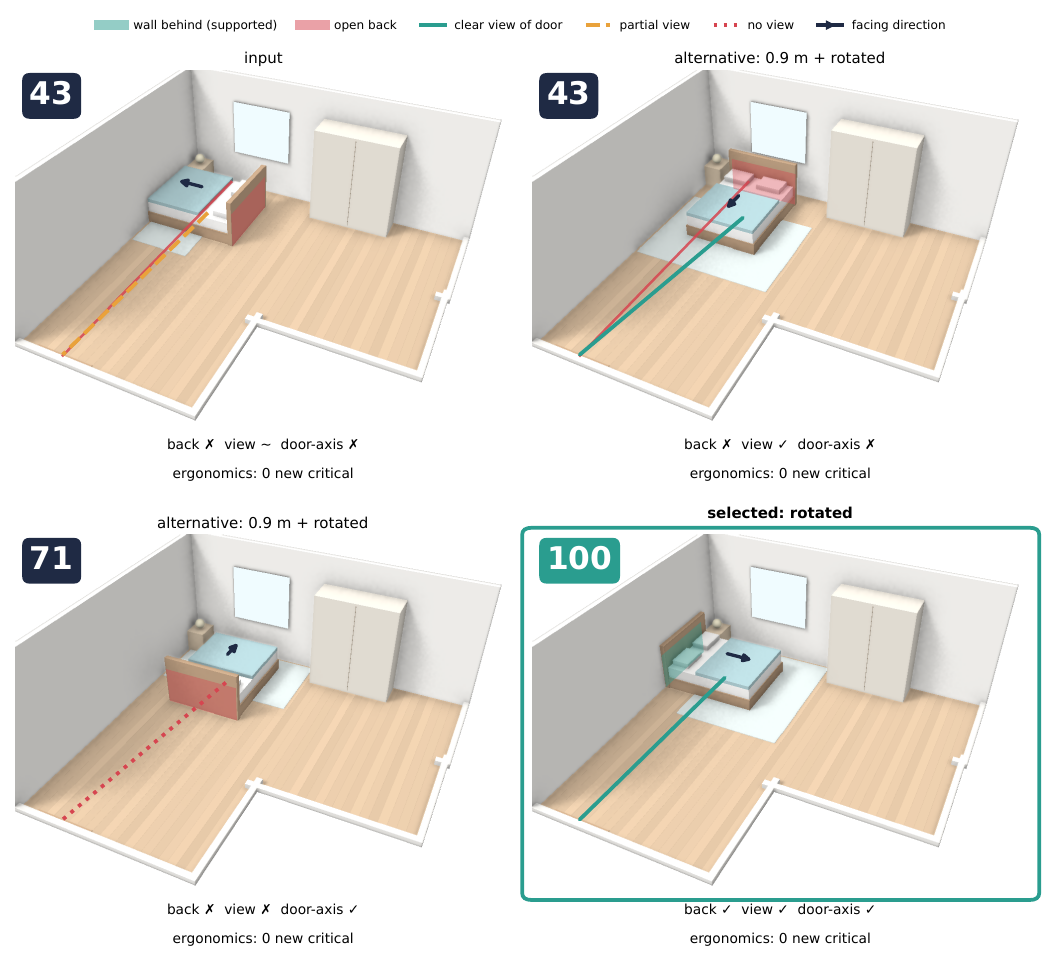}&
\includegraphics[viewport=249 29 505 211,clip,width=.485\linewidth]{figures/static_preference_bedroom_archive.pdf}\\[-2pt]
On the door axis & Clear of the door axis
\end{tabular}}
\caption{\textbf{Static preferences made visible.} Five-criterion rubric (Appendix~B): backing, visibility and door alignment. Results use the five-criterion definitions separately from HANFU.}
\label{fig:static-preferences}
\end{figure}

\paragraph{Declared preferences distinguish other arrangements.}
Under the five-criterion rubric (Appendix~B; version-specific semantics and results), image judgments score 37.5\% on eight shift pairs (Fig.~\ref{fig:preference-evidence}); this task is evaluated separately from Table~\ref{tab:static-ranking-main}. The ergonomics-first tandem resolves two critic ties. Sofa and bed examples illustrate backing, door visibility and axis alignment (Fig.~\ref{fig:static-preferences}).

\subsection{Household dynamics and apartment comparisons}
\paragraph{MAPPO captures persistent joint behavior.}
The recorded MAPPO actor completes an 18-goal day and a 126-goal weekly object relay with 21 successful handoffs. Another weekly routine completes 126/126 goals while traversing 175.95\,m. Event timelines show both residents contributing, including a stationary transfer partner. In a separate rich-corpus frozen comparison, MAPPO improves partial need coverage from 90 to 1,204 of 2,884 over its BC predecessor. These cohorts retain distinct interfaces and training protocols; Appendix~F connects their checkpoints, trajectories and iteration-level learning traces.

\begin{figure*}[t]\centering
\includegraphics[width=.94\textwidth]{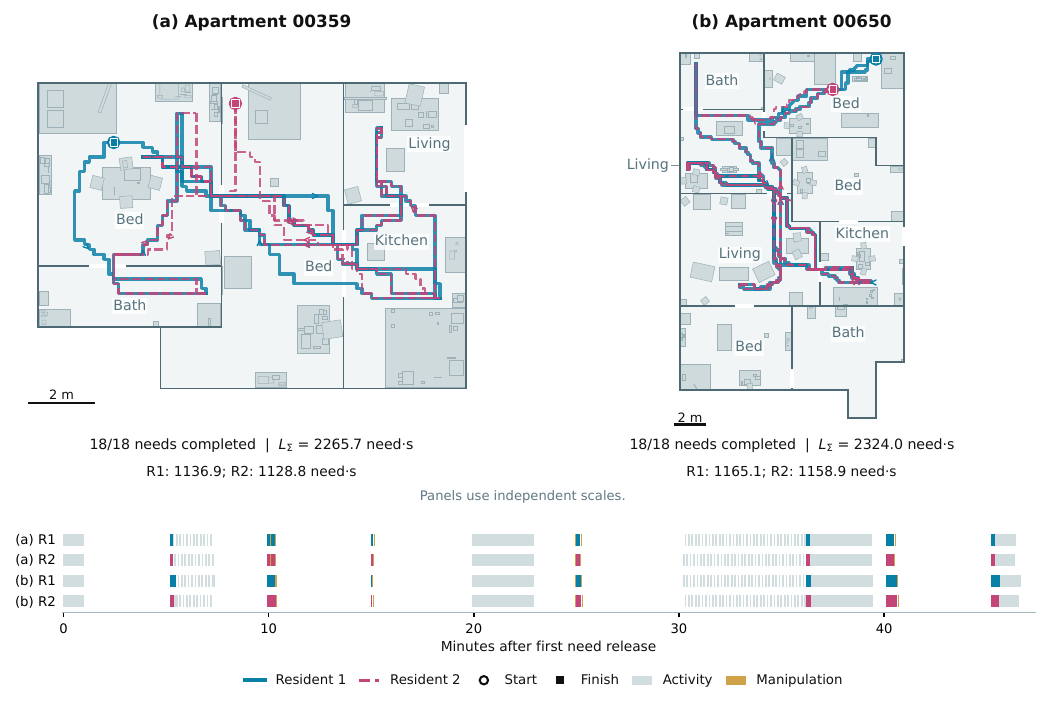}
\caption{\textbf{Routines across furnished homes.} Both reference executions fulfill all needs. Distances are in metres; repeated paths are drawn once. Timelines show manipulation and shared activity.}
\label{fig:whole-home-pair}
\end{figure*}

\paragraph{Routine competence enables interpretable comparisons.}
With weights fixed, current-state admissibility improves execution; routine composition additionally depends on learned behavior. Adaptation on complete demonstrations then enables nearly all matched whole-apartment comparisons, including cases with opposing resident costs. This matched checkpoint comparison combines additional training and loss balancing. Reference and finite temporal controllers fulfill all original needs in this cohort. During the illustrated evening, shorter travel outweighs additional resource waiting and benefits both residents (Fig.~\ref{fig:whole-home-pair}). During work at home, a slower resident reverses the reference aggregate preference while exposing a personal tradeoff. Temporal search removes that reversal. Thus the ordering is conditional on behavior and scheduling; Appendix~F preserves completion coverage and individual costs.

\begin{figure*}[t]\centering
\includegraphics[width=\textwidth]{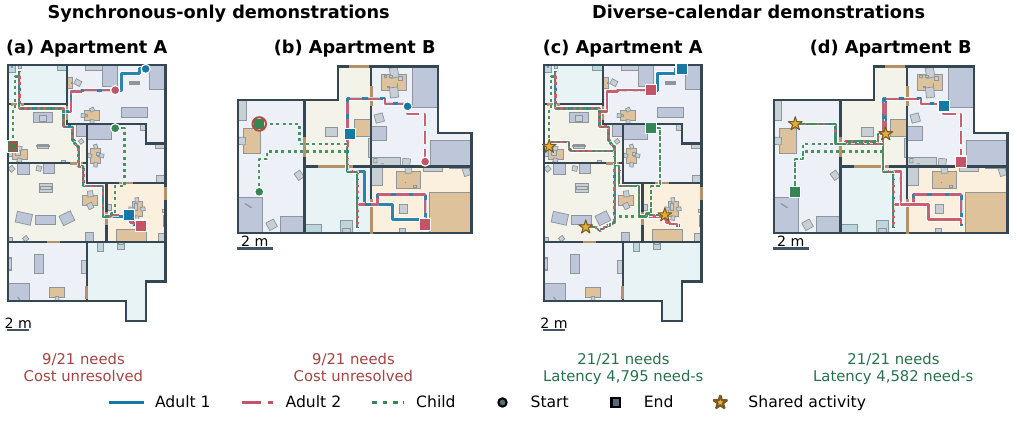}
\caption{\textbf{Calendar-diverse training enables complete family comparison.} For matched birthday routines, the synchronous-only actor stalls; diverse-calendar training completes both homes. Colors identify residents; stars mark shared activities. Audited routes are drawn once per distinct navigation path.}
\label{fig:family_calendar_routes}\end{figure*}

\begin{figure}[t]\centering
\includegraphics[width=\linewidth]{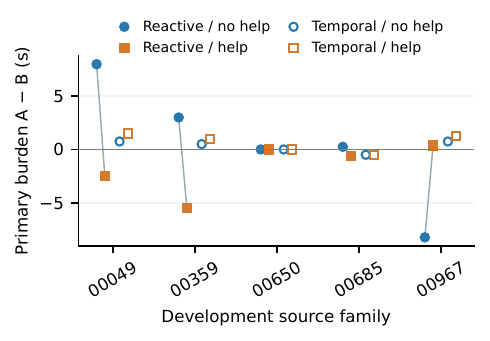}
\caption{\textbf{Assistance and scheduling.} Finite scheduling removes reactive reversals. Positive A--B favors B; points are individual development observations.}
\label{fig:help-rankings}\end{figure}

\paragraph{Shared resources and waiting explain interaction effects.}
Corrected-clock joint and solo controls fulfill all needs. With requests fixed, the interaction cost in Eq.~\ref{eq:interaction-cost} is positive under synchrony, vanishes as requests separate, and is zero with private resources. The mobility intervention preserves furnishing order in this control. Allowing one resident to deliver the other's object reverses reactive rankings in most families, yet interaction cost remains positive. Finite scheduling removes the primary-cost reversals; including voluntary waiting restores some of them (Fig.~\ref{fig:help-rankings}). Assistance changes furnishing preference through behavior and scheduling.

\paragraph{Learned behavior adds variation beyond temporal controls.}
Frozen actors have different completion coverage on matched joint and solo conditions. Among complete outcomes, two actors show the same local profile-dependent reversal; it persists for one frozen actor across additional request draws. The asynchronous contrast is less stable. Priority search matches reference costs, whereas delay search trades forced for voluntary waiting and removes the primary-cost interaction. An input-only route proxy explains optimized kitchen cost; learned execution retains nonzero synchronization effects in some complete comparisons. These controls separate behavior from geometry and accounting. Appendix~F attributes these contrasts to the observed actions and waiting mechanisms.

\subsection{Complete family routines across calendars}

Two adults and a child undertake after-work evenings, birthday preparation and celebrations. Six numerical profiles vary mobility, reach and assistance. Synchronous-trained actors lose coverage when requests shift, repeatedly selecting the child's table task without completing assistance. The traces locate the failure in coordinating assistance.

In the volume-changing development study, at matched update budgets, the larger diverse-calendar dataset restores all asynchronous routines (Table~\ref{tab:family_calendar}). Every development pair benefits all three residents, hence any nonnegative weighting with positive total weight agrees. This contrast changes both diversity and dataset size.

Without further fitting, both diverse-calendar actors complete all routines on two additional input-admitted TRAIN-pool homes. Synchronous-only controls again lose shifted-routine coverage. Most complete transfer comparisons involve opposing resident costs, unlike the unanimous development preferences. Successful transfer exposes these personal tradeoffs.
\begin{table}[t]\centering\small
\caption{Complete family scenarios and matched contexts. Each calendar: 72 scenarios ($2$ homes $\times3$ routines $\times6$ profiles $\times2$ request draws). The 72 contexts reuse one home-pair across both calendars. Parentheses index two paired training seeds; transfer rows hold for each seed.}
\label{tab:family_calendar}
\begin{tabular}{lccc}\toprule
Demonstrations & Sync. & Shifted & Contexts\\\midrule
\multicolumn{4}{l}{Development: 2 homes}\\
Synchronous only (1) &72/72&36/72&54/72\\
Synchronous only (2) &72/72&42/72&54/72\\
Diverse calendars (1) &72/72&\textbf{72/72}&\textbf{72/72}\\
Diverse calendars (2) &72/72&\textbf{72/72}&\textbf{72/72}\\\midrule
\multicolumn{4}{l}{Unfitted TRAIN-pool: 2 homes (each seed)}\\
Synchronous only &72/72&36/72&54/72\\
Diverse calendars &72/72&\textbf{72/72}&\textbf{72/72}\\\bottomrule
\end{tabular}\end{table}

On the development home-pair, both diverse-calendar actors agree in 72/72 matched contexts (Appendix~F); broader executor stability remains open. An equal-volume study varies calendar diversity across four initializations. Gains coexist with control-only successes. At new 60/330-second shifts, diverse actors complete 240/288/288/216 of 288 scenarios versus 144/156/255/210 for synchronous-only actors. These are temporal tests on the same homes (Appendix~G).

\paragraph{Task-covered furnishing examples.}
A directional wardrobe adds a 5.05\,m access detour despite tied static scores. Two further, fully completed case studies change both spatial and preference scores. A living-room rearrangement worsens both, yet shortens a route by 2.55\,m; its household cost preference reverses with resident speed. Audited trajectories and personal costs explain what each assessment adds on the same furnishings (Appendix~H).

\section{Discussion and Conclusion}
CoHABIT evaluates how furnished environments serve a household through complementary spatial criteria and everyday routines. Orientation and declared preferences distinguish arrangements left tied by footprint checks; shared routines reveal waiting, assistance and personal tradeoffs. The controlled examples connect these perspectives on identical furnishings: an access detour changes routine cost despite tied static scores, while a shorter route can coexist with worse ergonomic or preference scores. Resident capabilities and scheduling can also reverse an aggregate preference (Appendix~H).

Independent need verification grounds these comparisons in completed activities, while task coverage identifies which furnishings a routine actually tests. The resulting assessment supports reproducible, protocol-matched choices that explain both why an alternative is preferred and who benefits, retaining separate criteria and personal outcomes. Future work will broaden routines, calibrate household profiles and preferences, and extend the model to physical co-presence during movement (Appendix~D)\label{main:body-end}.

\clearpage
\section*{LLM Usage Statement}
OpenAI Codex and ChatGPT assisted with ideation, code implementation and review, experiment orchestration, analysis, manuscript drafting, and figure preparation. Conceptual AI renderings are labeled; measured figures derive from recorded experimental artifacts. The authors remain responsible for the scientific claims and submitted content.

\nocite{agarwal2021,foerster2018coma,gae2016,ippo2020,mask2022,ppo2017,rashid2018qmix,ross2011dagger,sutton1999options,xiong2020layernorm}

{\small\bibliographystyle{ieeenat_fullname}\bibliography{references}}
\clearpage
\input{appendix}
\end{document}